\documentclass[runningheads]{llncs}
\usepackage[T1]{fontenc}
\usepackage{graphicx}
\usepackage{booktabs}
\usepackage[misc]{ifsym}
\newcommand{\corr}{(\Letter)}
\usepackage{graphicx} 
\usepackage{hyperref}
\usepackage{amsmath} 
\usepackage{xcolor}
\usepackage{tabularx}
\usepackage{pifont}
\usepackage{multirow}
\usepackage{orcidlink}
\usepackage{amssymb}
\usepackage{amsfonts}
\usepackage[square, numbers]{natbib}

\newcommand{\libname}{Standardized Process Intelligence Comparison Engine}
\newcommand{\libnameshort}{SPICE }

\begin{document}

\title{Towards Reproducibility in Predictive Process Mining: \ \libnameshort --- A Deep Learning Library}

\titlerunning{\libnameshort --- A Deep Learning Library}


\author{Oliver Stritzel\inst{1}\orcidlink{0000-0003-1772-7428} \corr \and
 Nick Hühnerbein\inst{1}\orcidlink{0009-0009-6177-517X} \and
 Simon Rauch\inst{1}\orcidlink{0009-0008-4669-0656} \and
 Itzel Zarate\inst{1}\orcidlink{0009-0003-2378-9956} \and
 Lukas Fleischmann\inst{1} \and
 Moike Buck\inst{1} \and
 Attila Lischka\inst{1,2}\orcidlink{0009-0003-1371-5996} \and
 Christian Frey\inst{1,3}
 }


\authorrunning{O. Stritzel et al.}


\institute{Fraunhofer IIS, Nuremberg, Germany \email{\{oliver.stritzel,nick.huehnerbein,simon.rauch,itzel.zarate.zarate,\newline lukas.fleischmann,moike.buck\}@iis.fraunhofer.de}
 \and
 Chalmers University of Technology, Gothenburg, Sweden \email{lischka@chalmers.se}
 \and
 University of Technology Nuremberg, Nuremberg, Germany \email{christian.frey@utn.de}
 }

\maketitle              

\begin{abstract}
    In recent years, Predictive Process Mining (PPM) techniques based on artificial neural networks have evolved as a method for monitoring the future behavior of unfolding business processes and predicting Key Performance Indicators (KPIs).
    However, many PPM approaches lack reproducibility, transparency in decision-making, and the flexibility to incorporate new datasets or enable benchmarking, making meaningful comparisons across implementations difficult.
    In this paper, we propose \textit{\libnameshort}(\libname), a Python framework that reimplements three popular, existing baseline deep-learning-based models for PPM in PyTorch, while designing a common base framework with rigorous configurability to enable reproducible and robust comparison of past and future modelling approaches. We compare \libnameshort to original reported metrics and with balance-aware metrics on 11 openly available datasets.

\keywords{Predictive Process Mining  \and Machine Learning \and Deep Learning \and Reproducibility.}
\end{abstract}

\section{Introduction}

The rapid digitalization of business processes has enabled organizations to track operational events through digital traces. Process Mining (PM) provides post-hoc analysis of event logs, while Predictive Process Mining (PPM) extends this to proactive decision-making through e.g. suffix prediction, remaining time estimation, and outcome prediction for business process instances.

The sequential nature of process data has motivated Natural Language Processing (NLP)-inspired approaches --- particularly Long Short-Term Memory (LSTM) \cite{lstm} and Transformer-based models \cite{transformer} --- yielding numerous PPM publications \cite{Tax2017, camargo_learning_2019, 
processtransformer, ramamaneiro2021deeplearningpredictivebusiness}. However, unlike 
computer vision \cite{lecun2010mnist, deng2009imagenet, krizhevsky2009learningCIFAR, 
lin2014microsoftCOCO} or NLP \cite{NguyenRSGTMD16MSMARCO, wang2019glue}, prominent PPM datasets like the BPI Challenges lack predefined splits, leading to diverging preprocessing strategies and results that are difficult to reproduce or compare. This reproducibility crisis \cite{baker2016reproducibility, kapoor2023reproducibilitycrisis} persists despite efforts such as \cite{ramamaneiro2021deeplearningpredictivebusiness}, which itself inadvertently propagated methodological flaws from the pioneering works 
\cite{Tax2017, camargo_learning_2019, processtransformer} it benchmarks.

This work introduces \libnameshort --- \textit{\libname}, an open-source\footnote{Source code available at: https://gitlab.cc-asp.fraunhofer.de/iis-scs-a-publications/spice} Python library addressing three major pain points in current PPM literature:
\begin{enumerate}
    \item Reimplementation of prominent baseline models in PyTorch \cite{pytorch} for \textbf{Next Activity}, \textbf{Next Timestamp}, \textbf{Remaining Time}, and \textbf{Suffix} prediction.
    \item Standardization of data splitting, preprocessing, and removing experimental design flaws to enable fair comparison.
    \item Incorporation of high modularization to ensure extensibility to future works and improving existing architectures.
\end{enumerate}

\begin{figure}
    \centering
    \includegraphics[width=0.9\linewidth]{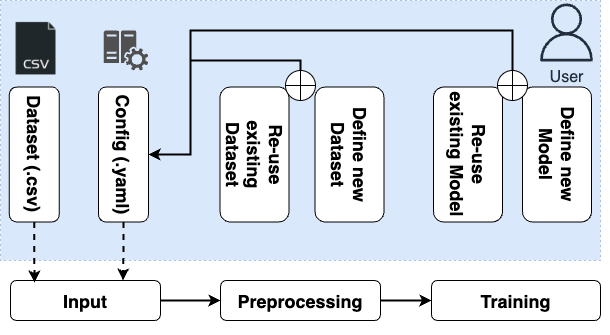}
    \caption{\libnameshort Design Overview --- Users can define new Models or Datasets or re-use existing ones. Configurations are combined in files which trigger one complete training process.}
    \label{fig:workflow}
\end{figure}

A design overview of \libnameshort is visualized in \autoref{fig:workflow}.

\section{Definitions \& Related Work}
This section begins with a formal definition of PPM, establishing the prediction tasks of interest and their corresponding optimization formulations. We then review three deep learning architectures that have significantly advanced the field, highlighting their (then) novel contributions. Finally, we address common experimental design challenges, including data splitting, information leakage, and reproducibility issues, which critically undermine the reliability of (current) PPM research.

\subsection{Predictive Process Mining} \label{definitions}

Event logs form the foundation of both PM and PPM. These sequential datasets capture process executions in form of events and must, by definition \cite{wmpvdaalst2012process}, include three core elements: \texttt{(1)} a unique case ID, \texttt{(2)} an activity, and \texttt{(3)} a timestamp for ordering events. Event logs often contain additional contextual information --- such as resource assignments, machine states, or sensor data --- yielding rich, multivariate datasets. Traditional PM emphasizes post-hoc analysis through process discovery, enhancement, and conformance checking \cite{wmpvdaalst2012process}. PPM, however, leverages advances in machine learning (ML) to shift the focus towards predictive monitoring of ongoing processes.

Following \cite{ramamaneiro2021deeplearningpredictivebusiness}, an event log is defined as a set of traces $D$, where each trace $\sigma \in D$ is grouped by case ID and chronologically ordered. A trace can be partitioned at position $k$ into an event prefix $\mathrm{hd}_k(\sigma) = \langle e_1, \ldots, e_k \rangle$ and an event suffix $\mathrm{tl}_k(\sigma) = \langle e_{k+1}, \ldots, e_{|\sigma|} \rangle$, where $|\sigma|$ denotes the trace length and each event $e_j$ carries an activity $A_j$, a timestamp $T_j$, and a time delta $\Delta T_j = T_j - T_{j-1}$. The objective is to learn a function $\Omega$ that predicts process KPIs based on these event prefixes. In practice, rather than using the entire prefix, a subsequence of $n$ consecutive events --- commonly termed \textit{n-grams} --- is extracted via a rolling window approach (see \autoref{fig:preprocessing}). The prediction tasks are defined as follows:

\paragraph{Next Activity \& Timestamp Prediction:}
These tasks focus on predicting the first element of the suffix $\mathrm{tl}_k(\sigma)$. Next activity prediction minimizes the expected negative log-likelihood over the discrete distribution $p$ of activities (Eq.~\eqref{eq:activity}), whereas next timestamp prediction targets the time delta $\Delta T_{k+1} = T_{k+1} - T_k$ under a continuous density $f$ (Eq.~\eqref{eq:timestamp}). Both tasks can be extended to multistep prediction (MSP).

    \begin{equation}
        \underset{\theta}{\operatorname*{argmin}}~\mathrm{E}_{\sigma \sim D} \left[- \log p \bigl(A_{k+1} \mid \mathrm{hd}_k(\sigma); \theta \bigr) \right]
        \label{eq:activity}
    \end{equation}

    \begin{equation}
    \begin{split}
        \underset{\theta}{\operatorname*{argmin}}~\mathrm{E}_{\sigma \sim D} \left[- \log f \bigl( \Delta T_{k+1} \mid \mathrm{hd}_k(\sigma); \theta \bigr) \right].
        \label{eq:timestamp}
    \end{split}
    \end{equation}

\paragraph{Suffix Prediction:}
Beyond single or MSP, \libnameshort supports suffix predictions to generate full case continuations by recursively applying the model to its own predictions. Starting with prefix $\mathrm{hd}_k(\sigma)$, the predicted suffix $\langle \hat{e}_{k+1}, \ldots, \hat{e}_{|\sigma|} \rangle$ is generated until an end-of-case token $\langle \texttt{END} \rangle$ is reached:
\begin{equation}
\hat{e}_{j} \sim p\bigl(e \mid \mathrm{hd}_k(\sigma) \oplus \langle \hat{e}_{k+1}, \ldots, \hat{e}_{j-1} \rangle; \theta \bigr),
\end{equation}
where $\oplus$ denotes sequence concatenation. Formally, the task of autoregressive suffix prediction of activity sequences is specified by:
\begin{equation}
    \hat{S} = \Omega_{suf}\bigl(\mathrm{hd}_k(\sigma)\bigr) = \langle \hat{A}_{k+1}, \ldots, \hat{A}_{|\sigma|} \rangle, \quad
\end{equation}
where $\hat{A}_j = \arg \max_A p\bigl(A \mid \mathrm{hd}_{j-1}(\hat{\sigma}); \theta \bigr)$
and $\hat{\sigma} = \mathrm{hd}_k(\sigma) \oplus \hat{S}$. Here, $\hat{S}$ contains only the predicted activities; timestamps are predicted analogously via $f$ and may be appended to each $\hat{e}_j$ before the next step. Sampling-based alternatives to argmax decoding are also supported (see \autoref{sec:library}).

We additionally implement MSP variants that forecast multiple future events in fewer forward passes by estimating $p\bigl(A_{k+1}, \ldots, A_{k+m} \mid \mathrm{hd}_k(\sigma); \theta \bigr)$ in one forward pass and minimizing the joint negative log-likelihood. When $m$ equals the suffix length, the entire suffix is predicted in a single pass, reducing inference time and error propagation at the cost of increased model complexity. The same applies to timestamp differences.

\paragraph{Remaining Time Prediction:}
Remaining time prediction estimates the time until case completion given $\mathrm{hd}_k(\sigma)$. 
Two approaches are supported. Direct prediction estimates the remaining time in a single forward 
pass as $\hat{R}_k = f\bigl(\mathrm{hd}_k(\sigma); \theta \bigr)$. Autoregressive prediction 
instead iteratively forecasts temporal increments $\widehat{\Delta T}_{k+j}$ until a stopping 
criterion is met --- either a fixed horizon $j_{\max}$ or
$\langle \texttt{END} \rangle$ predicted by the activity head --- accumulating the remaining time 
estimate as:
\begin{equation}
    \hat{R}_k = \sum_{j=1}^{m} \widehat{\Delta T}_{k+j}, 
    \quad \text{where } m = \min\!\left(\min\bigl\{j \geq 1 : \hat{A}_{k+j} = \langle \texttt{END} \rangle\bigr\},\ j_{\max}\right).
\end{equation}
While autoregressive prediction naturally couples activity and time forecasting and adapts the 
horizon to the predicted case length, direct prediction is generally preferred due to error 
propagation in iterative approaches~\cite{roider2024assessing}.

\paragraph{Outcome Prediction:}
Outcome prediction targets the process outcome $O$, such as quality assessments or customer satisfaction. Given a prefix, the objective is to estimate $\hat{O}=\Omega_{Out}\bigl(\mathrm{hd}_k(\sigma)\bigr)$, where $\hat{O} \in \{0,1\}$ for binary outcomes (e.g., pass/fail) or $\hat{O} \in [\min(O), \max(O)]$ for continuous scores, corresponding to classification and regression tasks, respectively.

\subsection{Deep Learning in PPM} \label{dl_in_ppm}

The sequential nature of event logs has motivated the adoption of NLP-inspired deep learning approaches in PPM. \cite{Evermann_2017} pioneered this direction by applying LSTM networks directly to event sequences using dense embeddings, though constrained to categorical variables without requiring an explicit process model.

\cite{Tax2017} extended this foundation with multi-task LSTM models that simultaneously predict next activity and timestamps, combining one-hot encoding with time-based feature enrichment. \cite{camargo_learning_2019} advanced LSTM-based PPM by integrating embeddings with tailored architectures and sophisticated preprocessing, demonstrating improved suffix prediction capabilities through shared architectures and log-normalized time features.

Following the groundbreaking impact of Transformer architectures, \cite{islam2023comprehensivesurveyapplicationstransformers, transformer}, \cite{processtransformer} replaced sequential LSTM processing with self-attention mechanisms, enabling parallel computation and more effective long-range dependency capture. Recent work has further adopted Transformer-based architectures for PPM tasks \cite{hennig2025transformericpm, hennig2025transformerbpm, wuyts2024sutran}.

We focus on \cite{Tax2017, camargo_learning_2019, processtransformer} due to their high citation impact as de-facto benchmarks, accessible codebases, and representation of key methodological advances: multi-task learning, optimized LSTM architectures, and attention-based models. Future versions of \libnameshort will incorporate additional baselines.

\subsection{Experimental Design \& Common Mistakes} \label{common_mistake}

Rigorous experimental design is essential for credible ML evaluation. Poor design choices --- such as biased dataset selection, inconsistent preprocessing, or inappropriate metrics --- can yield misleading conclusions \cite{ramamaneiro2021deeplearningpredictivebusiness}. While \cite{ramamaneiro2021deeplearningpredictivebusiness} addressed issues like uneven data splits, there is still a lack of examination of critical implementation errors, information leakage, and unfair preprocessing in existing approaches.

\paragraph{Data Splitting:~} 
Data splitting divides available data into training, validation, and test subsets to enable model fitting, hyperparameter selection, and unbiased generalization estimation \cite{raschka2020modelevaluationmodelselection, shah2025optimization}. In PPM, splits can be performed by case ID \cite{Tax2017, camargo_learning_2019, processtransformer}, by time \cite{abb_discussion_2023}, or by combining both \cite{wuyts2024sutran} as shown in \autoref{tab:datasplit_options}. Temporal splits raise challenges when cases span multiple periods; ideally, only completed cases within each split should be included, though this is difficult when case durations exceed the data collection window. Unbalanced activity distributions may also warrant stratified splits.

\begin{table}[h!]
\centering
\begin{tabularx}{0.95\textwidth}{l|X|X|X|X}
    \textbf{Strategy}   & \textbf{Train}   & \textbf{Validation} & \textbf{Test} & \textbf{Useful when...} \\
    \hline
    Case-ID based      & Distinct cases A   & Distinct cases B      & Distinct cases C & proving model capability for sequence prediction \\
    \hline
    Time-based          & Earliest events  & Middle events       & Latest events & investigating effects of incomplete information  \\
    \hline
    Combination         & Earliest finished cases & Intermediate finished cases & Latest finished cases & testing robustness against data drift \\
    \hline
    Stratified & Distinct sequences A & Distinct sequences B & Distinct sequences C & mitigating overfit towards majority sequences \\
\end{tabularx}
\caption{Assignment logic for splits under each strategy. Combination splits involve both case-IDs and temporal criteria.}
\label{tab:datasplit_options}
\end{table}

\paragraph{Information Leakage:~}  
Information leakage occurs when training data contains information unavailable during inference, leading to inflated performance estimates \cite{kapoor2023reproducibilitycrisis, Del_Grosso_2023}. This includes identical samples across splits, temporal leakage, or improper preprocessing. While repeated subsequences across splits are expected, preventing training on future information is crucial \cite{abb_discussion_2023}. PPM experiments must define objectives beforehand and design splits that respect temporal and causal constraints.

\paragraph{Reproducibility:~} 
Setting random seeds is fundamental for reproducibility, as stochastic processes can substantially impact results \cite{semmelrock2025reproducibilitymachinelearningbasedresearch}. While full reproducibility across hardware configurations may not always be achievable\footnote{\href{https://docs.pytorch.org/docs/stable/notes/randomness.html}{https://docs.pytorch.org/docs/stable/notes/randomness.html}}, publishing seeds, model artifacts, and using standardized datasets helps mitigate these limitations \cite{semmelrock2025reproducibilitymachinelearningbasedresearch, kapoor2023reproducibilitycrisis}.

\paragraph{Evaluation Metrics:~}
While Accuracy is commonly used for next activity prediction, its shortcomings for imbalanced datasets are well-documented \cite{spelmen2018review, brodersen2010balancedacc}. We therefore additionally report Balanced Accuracy for multiclass predictions.

For temporal predictions, Mean Absolute Error (MAE) is commonly used. Notably, existing approaches \cite{camargo_learning_2019, Tax2017} lack positive activation functions (e.g., softplus, ReLU)\footnote{\href{https://github.com/verenich/ProcessSequencePrediction/blob/9f51f9a30574a71465bb394a7ed82eee0338d10d/code/train.py\#L322}{https://github.com/verenich/ProcessSequencePrediction/train.py\#L322}, \href{https://github.com/AdaptiveBProcess/GenerativeLSTM/blob/16eb184093381dd91d0b99493fe52be197cd0de5/models/model_shared_cat.py\#L120}{https://github.com/AdaptiveBProcess/GenerativeLSTM\#L120}}
, potentially yielding negative predictions --- particularly problematic when MAE masks such errors\footnote{\href{https://github.com/verenich/ProcessSequencePrediction/blob/9f51f9a30574a71465bb394a7ed82eee0338d10d/code/train.py\#L328}{https://github.com/verenich/ProcessSequencePrediction/train.py\#L328}, \href{https://github.com/AdaptiveBProcess/GenerativeLSTM/blob/16eb184093381dd91d0b99493fe52be197cd0de5/models/model_shared_cat.py\#L137}{https://github.com/AdaptiveBProcess/GenerativeLSTM\#L137}}
. We retain these implementations for reproducibility.

For suffix prediction, the Normalized Damerau-Levenshtein (DL) Similarity $sim_{DL}=1-\frac{d_{DL}}{\max(|\sigma|,|\hat{\sigma}|)}$ is a commonly used metric \cite{Tax2017, camargo_learning_2019, ramamaneiro2021deeplearningpredictivebusiness}. It penalizes ordering errors not as much as incorrectly predicted activities (substitution costs 1; insertion/deletion costs 2). Alternative metrics include BLEU score or Jaccard Similarity.

In the following section, we describe how we tackle the listed flaws in current literature in our standardized framework \libnameshort.

\section{\libnameshort --- \libname}\label{sec:library}

\libnameshort follows a modular architecture adhering to PyTorch's DataLoader and model separation. Each approach retains its original data preparation specifics, provided they do not violate basic experimental design principles. Ideally, preprocessing methods and model implementations can be interchanged independently, as illustrated in \autoref{fig:workflow}.

\begin{figure}
    \centering
    \includegraphics[width=1\linewidth]{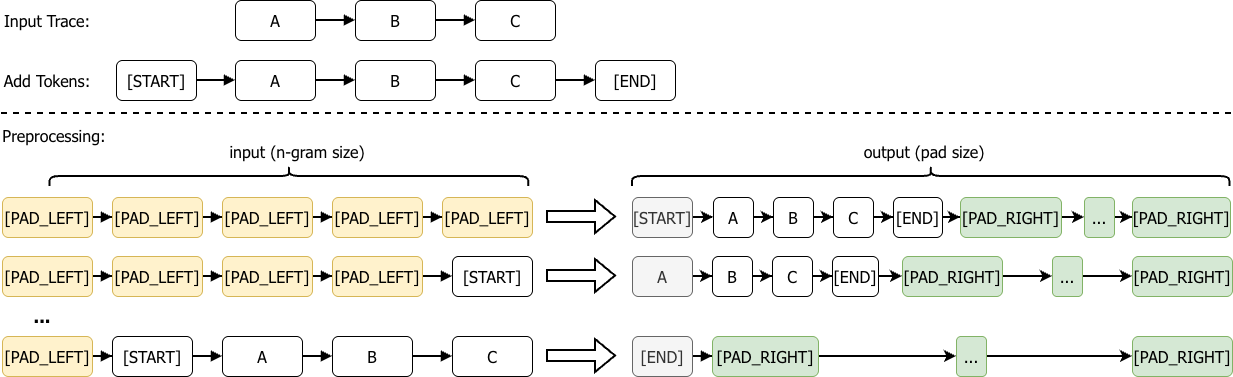}
    \caption{Preprocessing visualized: Raw traces are augmented with \texttt{START} and \texttt{END} tokens, and input-output pairs are padded to ensure equal sizes. A general preprocessing class creates these pairs; token encoding is implementation-specific. For next activity and timestamp prediction, only the first suffix element serves as target, though preprocessing remains identical, enabling multi-step prediction by design. Inputs and outputs can include time and resource features for multi-task models.}
    \label{fig:preprocessing}
\end{figure}

A central preprocessing class prepares input-output pairs for any dataset (see \autoref{fig:preprocessing}), with specific implementations handling only feature encoding (e.g., one-hot \cite{Tax2017} or embeddings \cite{camargo_learning_2019}). By default, the output is the remaining suffix with respective features; for single prediction tasks, only the first element of the output is selected as target, enabling a unified preprocessing pipeline across tasks.

The framework supports next activity and timestamp prediction using single-step or MSP strategies, extending to full suffix prediction via recursive generation or MSP. Remaining time estimation currently supports direct prediction only; iterative estimation via summed time increments, as well as outcome prediction, are planned for future versions.

Configuration management follows Hydra \cite{hydra} practices, with all experiment configurations available in the accompanying repository to ensure high reproducibility. \libnameshort incorporates recent advances in deep learning and MLOps, introducing numerous quality-of-life improvements over original implementations.

Multiple sampling strategies --- including Top-P sampling \cite{holtzman2020curiouscaseneuraltext}, Top-K sampling \cite{noarov2025foundationstopkdecodinglanguage}, and greedy sampling with temperature --- enable flexible control over model predictive behavior, moving beyond the argmax and classic random sampling used in original works. Experiments can be run in batches and logged to MLflow for comprehensive comparisons, with options to store trained models or output results to the console. New methods, datasets, samplers, optimizers, and other components can be easily integrated, and we invite community contributions.

By default, \libnameshort tracks Accuracy, Balanced Accuracy, and F1-score for next activity prediction; MAE, MSE, and RMSE for next timestamp and remaining time prediction; and $sim_{DL}$, BLEU Score, and Jaccard Similarity for suffix prediction.

\section{Implementation and Derivations}

This section presents the reimplementations within \libnameshort and deviations from original approaches. We focus on faithfully reproducing original methods while addressing identified shortcomings, such as information leakage and inappropriate normalization. Additionally, we outline improvements and modular design considerations that enhance reproducibility and extensibility.

\subsection{Camargo et al.} 

\paragraph{Pooling Role-Activity Pairs:~} The original implementation uses an algorithm by \cite{song_towards_2008}, a post-hoc analysis method for completed event logs lacking a fit/transform interface for unseen data. \cite{camargo_learning_2019} apply this procedure prior to data splitting, introducing information leakage by design. We thus removed it from our experimental setup. Future researchers should critically evaluate this approach, particularly when comparing against results from the original paper or \cite{ramamaneiro2021deeplearningpredictivebusiness}, which do not address this issue.

\paragraph{Pre-trained Embedding:~} The original paper trains embeddings on the entire dataset prior to splitting. We retain the pre-training procedure from \cite{camargo_learning_2019} but apply it only to the training split after data separation.

\paragraph{Batch Normalization between Recurrent Layers:~} Batch normalization \cite{ioffe2015batchnormalizationacceleratingdeep} in Recurrent Neural Networks (RNN) normalizes pre-activation hidden states across the batch dimension, making each hidden state dependent on other sequences through shared statistics. This violates RNN independence assumptions and introduces information leakage: identical sequences receive different normalizations across batches, destabilizing recurrent connections. Layer normalization \cite{ba2016layernormalization} instead computes statistics within each sequence independently, preserving sequential processing requirements. We therefore reimplement the multilayer LSTM from \cite{camargo_learning_2019} using layer normalization between RNN layers.

\subsection{Tax et al.}

\paragraph{Activity Encodings:} The DL-Similarity is designed for character-level string comparison. \cite{Tax2017} circumvented this by encoding activities as special characters --- a fundamentally flawed approach due to system-dependent encodings. Their implementation acknowledges anomalous similarity values on certain systems\footnote{\href{https://github.com/verenich/ProcessSequencePrediction/blob/9f51f9a30574a71465bb394a7ed82eee0338d10d/code/evaluate_suffix_and_remaining_time.py\#L278}{https://github.com/verenich/ProcessSequencePrediction\#L278}}, addressed by setting problematic results to zero rather than resolving the underlying issue. To ensure reliable calculations, we developed a native implementation operating directly on activity sequences, applied uniformly across all experiments.

\paragraph{Batch Normalization:} Analogous to \cite{camargo_learning_2019}, we apply layer normalization for reproducibility, with layerwise normalization also implemented.

\paragraph{Time Features:} The original paper mentions three time features, but the code applies five (including time since case start and a position index). Two features are scaled by means computed over the entire dataset before splitting, introducing information leakage. We restrict scaling to training data only. We also use time differences in days rather than hours, consistent with reported results. The original implementation inconsistently scales features (e.g., position index and time since last Sunday midnight remain unscaled), which we retain for comparability.

\subsection{Buksh et al.}\label{subsec:buksh}

\paragraph{Unweighted Accuracy Calculation:} The evaluation procedure calculates metrics for each prefix length $k$, but aggregates results using unweighted means of per-prefix scores. As sample sizes vary substantially across prefixes, this approach biases estimates toward prefixes with fewer cases (i.e., longer prefixes). A weighted aggregation reflecting actual sample distributions would provide more representative performance summaries. \cite{camargo_learning_2019} may exhibit the same issue\footnote{\href{https://github.com/AdaptiveBProcess/GenerativeLSTM/blob/16eb184093381dd91d0b99493fe52be197cd0de5/predict_suffix_full.py\#L110}{https://github.com/AdaptiveBProcess/GenerativeLSTM\#L110}}.

\paragraph{ProcessTransformer vs. Established Transformers:} The implementation uses global mean pooling followed by dense classification layers, which is incompatible with next token prediction in standard encoder-decoder \cite{transformer} and decoder-only transformers \cite{radford2018improving}. Both architectures preserve sequential structure through causal attention and position-wise vocabulary projections rather than collapsing via global pooling. For reproducibility, we reimplemented ProcessTransformer with its original design choices.

\section{Experimental Setup}

\paragraph{Datasets:~}
We evaluate on eleven benchmark datasets: BPI 2012, BPI 2013, BPI 2014, BPI 2015-1, BPI 2017, BPI 2019, BPI 2020d, BPI 2020i, Helpdesk, Hospital Billing, and Road Traffic. A key observation is that despite each paper comparing against prior work, they lack a common experimental design and use different dataset variations. 
We therefore exclude dataset derivatives (e.g., BPI 2012W, Helpdesk's anonymized version) to establish a clean baseline for past and future comparisons. For comparisons in \autoref{subsec:eval}, we report results for BPI 2012 and Helpdesk only. \autoref{subsec:fair-eval} provides plain and balanced accuracy for next activity prediction across all datasets.

\paragraph{Data Splitting:~}
The considered approaches employ diverging splitting strategies:
\begin{itemize}
    \item \cite{Tax2017} use a chronological $\frac{2}{3}$/$\frac{1}{3}$ split, but their line-based counting may divide cases and fails on non-chronologically ordered datasets.
    \item \cite{camargo_learning_2019} employ a 70/30 split ordered by case end timestamps, removing cases spanning both splits (combined split without validation set).
    \item \cite{processtransformer} describe a 64/16/20 chronological split, but their implementation does not handle case overlaps, effectively resulting in case-ID based splits.
\end{itemize}

We implement an 80/10/10 train-validation-test split by case-ID. While cross-validation would more rigorous, the recursive computational demands of suffix prediction \cite{camargo_learning_2019, ramamaneiro2021deeplearningpredictivebusiness} make it impractical for this study.

\paragraph{Preprocessing:~}
Standardized input processing follows \autoref{sec:library}, with n-gram sizes matching respective paper settings. For padding, instead of computing maximum trace length $l_{max} = \max_{\sigma \in D}(|\sigma|)$ over the complete dataset, we use pad size $p = 2^s$ where $s$ is the smallest integer such that $2^s > l_{\max}$, computed on training data only to avoid information leakage..

\paragraph{Models:~}
We ported models from \cite{Tax2017}, \cite{camargo_learning_2019}, and \cite{processtransformer} to PyTorch, staying as close as possible to the original implementations
\footnote{ \href{https://github.com/AdaptiveBProcess/GenerativeLSTM}{https://github.com/AdaptiveBProcess/GenerativeLSTM}, 
\href{https://github.com/Zaharah/processtransformer}{https://github.com/Zaharah/processtransformer}
}.
Hyperparameters follow authors' best reported configurations; when multiple combinations yielded identical scores, we conducted experiments to determine optimal settings.

\paragraph{Training:~}
We use cross-entropy loss for next activity prediction and MAE for next timestamp prediction, except for \cite{processtransformer} which employs log-cosh loss $\frac{1}{b}\sum^b_i\log($ $\cosh(\hat{y_i}-y_i))$. Models train for 100 epochs with early stopping (patience=10), saving the best model based on validation loss.

\paragraph{Sampling:~}
Sampling strategies diverge across approaches: \cite{camargo_learning_2019} found random sampling superior to argmax (used in our experiments), while \cite{Tax2017} and \cite{processtransformer} employ argmax sampling based on repository inspection.

\paragraph{Evaluation Metrics:~}\label{evaluation}
For comparability with original papers, we report accuracy for next activity prediction and MAE for next timestamp prediction, alongside balanced accuracy to address label imbalance.

\section{Evaluation} \label{sec:evaluation}
This section compares our reimplementations against original reported results. Temporal metrics are aligned to days, and suffix prediction quality is measured via $sim_{DL}$; balanced accuracy results appear in \autoref{subsec:fair-eval}. Suffix prediction scores may be systematically lower due to increased pad sizes, which allow more room for hallucination errors.

\subsection{Evaluation \& Metric Comparison}\label{subsec:eval}

\paragraph{Camargo et al.:~}
Results for \cite{camargo_learning_2019} are shown in \autoref{tab:camargo_results}. Next activity prediction improved overall, possibly due to our smaller test split (10\%). No next timestamp metrics were reported originally.

For remaining time, the original paper uses an indirect approach: predicting each suffix activity with corresponding timestamps, then calculating remaining time from the last predicted timestamp. We predict remaining time directly, with improvements consistent with \cite{ramamaneiro2021deeplearningpredictivebusiness} findings that direct prediction outperforms iterative approaches.

Suffix prediction metrics decreased significantly, likely due to preprocessing changes. The original implementation never handled unknown tokens since activity-resource pooling was performed on the complete dataset. Other reviews \cite{ramamaneiro2021deeplearningpredictivebusiness} also failed to reproduce original metrics. The use of random sampling for suffix generation introduces additional reproducibility challenges, suggesting original scores may be optimistic.

\begin{table}[h]
\centering
\begin{tabular}{l|l|l|l|l|l|l|l|l}
        & \multicolumn{2}{l}{Next Activity $^a$} \vline & \multicolumn{2}{l}{Next Timestamp $^e$} \vline & \multicolumn{2}{l}{Suffix $^d$} \vline & \multicolumn{2}{l}{Remaining Time $^e$} \\ \hline 
Dataset & Original          & Ours          & Original        & Ours        & Original       & Ours      & Original $i$          & Ours          \\ \hline \hline
BPI 2012   & 0.786 & \textbf{0.869} & --- & \textbf{0.371} & \textbf{0.632} & 0.390 & 8 $*$ & \textbf{7.267} \\ \hline
Helpdesk   & 0.789 & \textbf{0.865} & --- & \textbf{3.096} & \textbf{0.917} & 0.698 & 6.1   & \textbf{5.962}
\end{tabular}
\caption{Evaluation Metrics of \cite{camargo_learning_2019} and our results. $*$ Compared to BPI 2012W as this is the only metric available from the original paper. $i$ Metrics approximated from bar plots. $^a$ in Accuracy; $^e$ in MAE (days); $^d$ in $sim_{DL}$}
\label{tab:camargo_results}
\end{table}    

\paragraph{Tax et al.:~}
\autoref{tab:tax_results} compares our results with \cite{Tax2017}. Next activity prediction shows slight improvements, again possibly due to our 10\% test split. Next timestamp results are comparable after converting from hours to days; differences for BPI 2012 may stem from our use of the original dataset versus the derived version.

Suffix prediction used argmax sampling, significantly affecting outcomes. For BPI 2012, greedy sampling yielded lower similarity scores; random sampling improved results to $sim_{DL} = 0.391$. \cite{ramamaneiro2021deeplearningpredictivebusiness} also reported lower metrics ($sim_{DL}=0.1409$) without specifying their sampler. Complex logs like BPI 2012 benefit from random sampling for more diverse sequences, while simpler logs like Helpdesk perform better with greedy sampling due to unbalanced label distributions. This suggests that the use of flexible sampling strategies (Top-P, Top-K) may be beneficial.

\begin{table}[h!]
\centering
\begin{tabular}{l|l|l|l|l|l|l|l|l}
        & \multicolumn{2}{l}{Next Activity $^a$} \vline & \multicolumn{2}{l}{Next Timestamp $^e$} \vline & \multicolumn{2}{l}{Suffix $^d$} \vline & \multicolumn{2}{l}{Remaining Time $^e$} \\ \hline
Dataset & Original $i$         & Ours          & Original $ii$        & Ours        & Original       & Ours  & Original $iii$     & Ours          \\ \hline \hline
BPI 2012 $*$  & 0.760 & \textbf{0.877} & 1.45  & \textbf{0.313} & \textbf{0.353} & 0.194 & \textbf{5.8} & 7.298 \\ \hline
Helpdesk      & 0.712 & \textbf{0.870} & \textbf{2.11}  & 2.825 & 0.767 & \textbf{0.898} & \textbf{4.2} & 6.03
\end{tabular}
\caption{Evaluation Metrics of \cite{Tax2017} and our results. $*$ \cite{Tax2017} report metrics for BPI 2012W; we used standard BPI 2012. $i$ Model: $layers=2$, $layers_{shared}=1$, $ngram=10$. $ii$ Model: $layers=2$, $layers_{shared}=1$. $iii$ Metrics approximated from bar plots. $^a$ in Accuracy; $^e$ in MAE (days); $^d$ in $sim_{DL}$}
\label{tab:tax_results}
\end{table}

\paragraph{Buksh et al.:~}
Results for \cite{processtransformer} in \autoref{tab:transformer_results} show slight differences for next activity prediction. No suffix metrics were reported originally; our results suggest the transformer performs best here due to its capacity for analyzing long-range dependencies.

Next timestamp and remaining time results are worse, possibly linked to log-cosh loss being less penalizing on errors, particularly for smaller datasets like Helpdesk. No reference comparison is available as \cite{ramamaneiro2021deeplearningpredictivebusiness} did not include this model. As noted in \autoref{subsec:buksh}, accuracy computation is unweighted, which also affects temporal predictions.

\begin{table}[h!]
\centering
\begin{tabular}{l|l|l|l|l|l|l|l|l}
        & \multicolumn{2}{l}{Next Activity $^a$} \vline & \multicolumn{2}{l}{Next Timestamp $^e$} \vline & \multicolumn{2}{l}{Suffix $^d$} \vline & \multicolumn{2}{l}{Remaining Time $^e$} \\ \hline
Dataset & Original          & Ours          & Original        & Ours        & Original       & Ours      & Original           & Ours          \\ \hline \hline
BPI 2012      & \textbf{0.852} & 0.830  & \textbf{0.250} & 0.658 &  ---  & \textbf{0.403} & \textbf{4.6}  & 7.8 \\ \hline
Helpdesk      & \textbf{0.856} & 0.837 & \textbf{2.980} & 9.202 &  ---  & \textbf{0.856} & \textbf{3.72} & 5.888 
\end{tabular}
\caption{Evaluation Metrics of \cite{processtransformer} and our results. Original paper did not include suffix prediction experiments. $^a$ in Accuracy; $^e$ in MAE (days); $^d$ in $sim_{DL}$}
\label{tab:transformer_results}
\end{table}

\subsection{Comparison with Balance-Aware Metrics}\label{subsec:fair-eval}

We provide results for all datasets in \autoref{tab:bal_acc}. For datasets not used in original papers, hyperparameters were selected from the most similar dataset based on characteristics: For \cite{processtransformer}, default settings were retained over all runs. For \cite{camargo_learning_2019} and \cite{Tax2017}, BPI 2020d, BPI 2020i, and Road Traffic used Helpdesk parameters (similar activity counts and case lengths), while for Hospital, BPI 2013 settings were used (BPI 2012 for Tax). Additionally, \cite{Tax2017} parameters from BPI 2012 were applied to BPI 2013 and BPI 2015-1.

\autoref{tab:bal_acc} reports next activity prediction metrics comparing balanced accuracy to plain accuracy. As expected, all models perform worse on balanced accuracy. The performance drop for highly imbalanced datasets like Helpdesk is particularly concerning, suggesting that models fail to capture deviations from standard process paths. This limitation is significant, as deviations from the norm often represent the most critical process instances, and actionable insights for these cases are arguably most beneficial. Consequently, models biased toward frequent process paths provide limited practical utility.

\begin{table}[h!]
\centering
\begin{tabular}{l|l|l|l|l|l|l}
        & \multicolumn{2}{l}{Camargo et al.} \vline & \multicolumn{2}{l}{Tax et al.} \vline & \multicolumn{2}{l}{Buksh et al.} \\ \hline
Dataset & Acc          & Bal. Acc          & Acc        & Bal. Acc        & Acc       & Bal. Acc           \\ \hline \hline
BPI 2012      & 0.869 & 0.655 & \textbf{0.877} & \textbf{0.692} & 0.830 & 0.614 \\ \hline
BPI 2013      & \textbf{0.801} & \textbf{0.662} & 0.764 & 0.611 & 0.726 & 0.499 \\ \hline
BPI 2015-1    & \textbf{0.540} & 0.344 & 0.533 & \textbf{0.364} & 0.480 & 0.261 \\ \hline
BPI 2020d     & \textbf{0.912} & \textbf{0.496} & \textbf{0.912} & \textbf{0.496} & 0.898 & 0.462 \\ \hline
BPI 2020i     & 0.859 & 0.480 & \textbf{0.887} & \textbf{0.517} & 0.869 & 0.491 \\ \hline
Helpdesk      & 0.865 & 0.398 & \textbf{0.870} & \textbf{0.408} & 0.837 & 0.350 \\ \hline
Hospital      & \textbf{0.942} & 0.584 & 0.926 & \textbf{0.598} & 0.844 & 0.464 \\ \hline
Road Traffic  & 0.879 & \textbf{0.694} & \textbf{0.882} & 0.698 & 0.853 & 0.667 \\ 
\end{tabular}
\caption{Accuracy and Balanced Accuracy on the test split. Balanced Accuracy decline is larger for datasets with more imbalanced activity distributions.}
\label{tab:bal_acc}
\end{table}

\section{Conclusion and Future Work}

This work critically outlines a number of deviations in experimental designs recent papers that are frequently used to benchmark new approaches and research ideas in the field of PPM. We have collected and aggregated opinions and findings that came up during researching the respective modelling approaches and discussed fundamental questions regarding fair data splitting, the use of class-imbalance-aware metrics, and fundamental design flaws. By that, we illustrate the shortcomings of the de-facto baseline models in PPM \cite{Tax2017, camargo_learning_2019, processtransformer}. In our opinion, future research should keep the deviations highlighted in this work in mind and refrain from building on top of these flawed designs as such.

Therefore, we provide a framework on which future research can build, enabling ablation studies, out-of-the-box comparison of newer methods with older ones, and the recreation of trustworthy baseline metrics using modern ML standards. We encourage researchers to reuse, modify, and extend our framework in the future. Furthermore, we believe that research in PPM should not stop at providing metrics but should also aim for practical impact by including evaluations with key-user studies to demonstrate real-world usefulness, while acknowledging that public process data are scarce and that finding key users for evaluations is challenging. Without such evidence, however, the risk of creating just another theoretical research bubble increases.

%
%
%
\bibliographystyle{splncs04}
\bibliography{lib}

@misc{ramamaneiro2021deeplearningpredictivebusiness,
      title={Deep Learning for Predictive Business Process Monitoring: Review and Benchmark}, 
      author={Efrén Rama-Maneiro and Juan C. Vidal and Manuel Lama},
      year={2021},
      eprint={2009.13251},
      archivePrefix={arXiv},
      primaryClass={cs.LG},
      url={https://arxiv.org/abs/2009.13251}, 
}

@inproceedings{camargo_learning_2019,
	address = {Berlin, Heidelberg},
	title = {Learning {Accurate} {LSTM} {Models} of {Business} {Processes}},
	isbn = {978-3-030-26618-9},
	url = {https://doi.org/10.1007/978-3-030-26619-6_19},
	doi = {10.1007/978-3-030-26619-6_19},
	urldate = {2023-10-04},
	booktitle = {Business {Process} {Management}: 17th {International} {Conference}, {BPM} 2019, {Vienna}, {Austria}, {September} 1–6, 2019, {Proceedings}},
	publisher = {Springer-Verlag},
	author = {Camargo, Manuel and Dumas, Marlon and González-Rojas, Oscar},
	month = sep,
	year = {2019},
	pages = {286--302},
}

@article{Evermann_2017,
   title={Predicting process behaviour using deep learning},
   volume={100},
   ISSN={0167-9236},
   url={http://dx.doi.org/10.1016/j.dss.2017.04.003},
   DOI={10.1016/j.dss.2017.04.003},
   journal={Decision Support Systems},
   publisher={Elsevier BV},
   author={Evermann, Joerg and Rehse, Jana-Rebecca and Fettke, Peter},
   year={2017},
   month=aug, pages={129–140} }

@misc{abb_discussion_2023,
	title = {A {Discussion} on {Generalization} in {Next}-{Activity} {Prediction}},
	url = {http://arxiv.org/abs/2309.09618},
	doi = {10.48550/arXiv.2309.09618},
	urldate = {2025-02-06},
	publisher = {arXiv},
	author = {Abb, Luka and Pfeiffer, Peter and Fettke, Peter and Rehse, Jana-Rebecca},
	month = sep,
	year = {2023},
	note = {arXiv:2309.09618 [cs]},
}

@article{song_towards_2008,
	title = {Towards comprehensive support for organizational mining},
	volume = {46},
	issn = {0167-9236},
	url = {https://www.sciencedirect.com/science/article/pii/S0167923608001280},
	doi = {10.1016/j.dss.2008.07.002},
	number = {1},
	urldate = {2025-05-30},
	journal = {Decision Support Systems},
	author = {Song, Minseok and van der Aalst, Wil M. P.},
	month = dec,
	year = {2008},
	pages = {300--317},
}

@article{ioffe2015batchnormalizationacceleratingdeep,
      title={Batch Normalization: Accelerating Deep Network Training by Reducing Internal Covariate Shift}, 
      author={Sergey Ioffe and Christian Szegedy},
      year={2015},
      eprint={1502.03167},
      archivePrefix={arXiv},
      primaryClass={cs.LG},
      url={https://arxiv.org/abs/1502.03167}, 
}

@misc{ba2016layernormalization,
      title={Layer Normalization}, 
      author={Jimmy Lei Ba and Jamie Ryan Kiros and Geoffrey E. Hinton},
      year={2016},
      eprint={1607.06450},
      archivePrefix={arXiv},
      primaryClass={stat.ML},
      url={https://arxiv.org/abs/1607.06450}, 
}

@inproceedings{Tax2017,
  title={Predictive business process monitoring with {LSTM} neural networks},
  author={Tax, Niek and Verenich, Ilya and La Rosa, Marcello and Dumas, Marlon},
  booktitle={Proceedings of the 29th International Conference on Advanced Information Systems Engineering},
  year={2017},
  pages={477--492},
  publisher={Springer}
}

@inproceedings{deng2009imagenet,
  title={Imagenet: A large-scale hierarchical image database},
  author={Deng, Jia and Dong, Wei and Socher, Richard and Li, Li-Jia and Li, Kai and Fei-Fei, Li},
  booktitle={2009 IEEE conference on computer vision and pattern recognition},
  pages={248--255},
  year={2009},
  organization={Ieee}
}

@inproceedings{lin2014microsoftCOCO,
  title={Microsoft coco: Common objects in context},
  author={Lin, Tsung-Yi and Maire, Michael and Belongie, Serge and Hays, James and Perona, Pietro and Ramanan, Deva and Doll{\'a}r, Piotr and Zitnick, C Lawrence},
  booktitle={European conference on computer vision},
  pages={740--755},
  year={2014},
  organization={Springer}
}

@article{krizhevsky2009learningCIFAR,
  title={Learning multiple layers of features from tiny images},
  author={Krizhevsky, Alex and Hinton, Geoffrey and others},
  year={2009},
  publisher={Toronto, ON, Canada}
}

@inproceedings{wang2019glue,
  title={{GLUE}: A Multi-Task Benchmark and Analysis Platform for Natural Language Understanding},
  author={Wang, Alex and Singh, Amanpreet and Michael, Julian and Hill, Felix and Levy, Omer and Bowman, Samuel R.},
  note={In the Proceedings of ICLR.},
  year={2019}
}

@article{NguyenRSGTMD16MSMARCO,
  author    = {Tri Nguyen and
               Mir Rosenberg and
               Xia Song and
               Jianfeng Gao and
               Saurabh Tiwary and
               Rangan Majumder and
               Li Deng},
  title     = {{MS} {MARCO:} {A} Human Generated MAchine Reading COmprehension Dataset},
  journal   = {CoRR},
  volume    = {abs/1611.09268},
  year      = {2016},
  url       = {http://arxiv.org/abs/1611.09268},
  archivePrefix = {arXiv},
  eprint    = {1611.09268},
  bibsource = {dblp computer science bibliography, https://dblp.org}
}

@article{radford2018improving,
  title={Improving language understanding by generative pre-training},
  author={Radford, Alec and Narasimhan, Karthik and Salimans, Tim and Sutskever, Ilya and others},
  year={2018},
  publisher={San Francisco, CA, USA}
}

@Misc{hydra,
  author =       {Omry Yadan},
  title =        {Hydra - A framework for elegantly configuring complex applications},
  howpublished = {Github},
  year =         {2019},
  url =          {https://github.com/facebookresearch/hydra}
}

@article{roider2024assessing,
  title={Assessing the performance of remaining time prediction methods for business processes},
  author={Roider, Johannes and Nguyen, An and Zanca, Dario and Eskofier, Bjoern M},
  journal={IEEE Access},
  year={2024},
  publisher={IEEE}
}

@misc{pytorch,
      title={PyTorch: An Imperative Style, High-Performance Deep Learning Library}, 
      author={Adam Paszke and Sam Gross and Francisco Massa and Adam Lerer and James Bradbury and Gregory Chanan and Trevor Killeen and Zeming Lin and Natalia Gimelshein and Luca Antiga and Alban Desmaison and Andreas Köpf and Edward Yang and Zach DeVito and Martin Raison and Alykhan Tejani and Sasank Chilamkurthy and Benoit Steiner and Lu Fang and Junjie Bai and Soumith Chintala},
      year={2019},
      eprint={1912.01703},
      archivePrefix={arXiv},
      primaryClass={cs.LG},
      url={https://arxiv.org/abs/1912.01703}, 
}

@misc{noarov2025foundationstopkdecodinglanguage,
      title={Foundations of Top-$k$ Decoding For Language Models}, 
      author={Georgy Noarov and Soham Mallick and Tao Wang and Sunay Joshi and Yan Sun and Yangxinyu Xie and Mengxin Yu and Edgar Dobriban},
      year={2025},
      eprint={2505.19371},
      archivePrefix={arXiv},
      primaryClass={cs.AI},
      url={https://arxiv.org/abs/2505.19371}, 
}

@misc{holtzman2020curiouscaseneuraltext,
      title={The Curious Case of Neural Text Degeneration}, 
      author={Ari Holtzman and Jan Buys and Li Du and Maxwell Forbes and Yejin Choi},
      year={2020},
      eprint={1904.09751},
      archivePrefix={arXiv},
      primaryClass={cs.CL},
      url={https://arxiv.org/abs/1904.09751}, 
}

@article{shah2025optimization,
  title={Optimization Of Data Splitting Methods For Machine Learning},
  author={Shah, Krutik and Shah, Shubh and Shah, Vatsal and Godbole, Prof and others},
  journal={Optimization Of Data Splitting Methods For Machine Learning (March 23, 2025)},
  year={2025}
}

@misc{raschka2020modelevaluationmodelselection,
      title={Model Evaluation, Model Selection, and Algorithm Selection in Machine Learning}, 
      author={Sebastian Raschka},
      year={2020},
      eprint={1811.12808},
      archivePrefix={arXiv},
      primaryClass={cs.LG},
      url={https://arxiv.org/abs/1811.12808}, 
}

@misc{semmelrock2025reproducibilitymachinelearningbasedresearch,
      title={Reproducibility in Machine Learning-based Research: Overview, Barriers and Drivers}, 
      author={Harald Semmelrock and Tony Ross-Hellauer and Simone Kopeinik and Dieter Theiler and Armin Haberl and Stefan Thalmann and Dominik Kowald},
      year={2025},
      eprint={2406.14325},
      archivePrefix={arXiv},
      primaryClass={cs.SE},
      url={https://arxiv.org/abs/2406.14325}, 
}

@article{Del_Grosso_2023,
   title={Bounding information leakage in machine learning},
   volume={534},
   ISSN={0925-2312},
   url={http://dx.doi.org/10.1016/j.neucom.2023.02.058},
   DOI={10.1016/j.neucom.2023.02.058},
   journal={Neurocomputing},
   publisher={Elsevier BV},
   author={Del Grosso, Ganesh and Pichler, Georg and Palamidessi, Catuscia and Piantanida, Pablo},
   year={2023},
   month=may, pages={1–17} }

@misc{islam2023comprehensivesurveyapplicationstransformers,
      title={A Comprehensive Survey on Applications of Transformers for Deep Learning Tasks}, 
      author={Saidul Islam and Hanae Elmekki and Ahmed Elsebai and Jamal Bentahar and Najat Drawel and Gaith Rjoub and Witold Pedrycz},
      year={2023},
      eprint={2306.07303},
      archivePrefix={arXiv},
      primaryClass={cs.LG},
      url={https://arxiv.org/abs/2306.07303}, 
}

@article{lstm,
  author={Hochreiter, Sepp and Schmidhuber, Jürgen},
  journal={Neural Computation}, 
  title={Long Short-Term Memory}, 
  year={1997},
  volume={9},
  number={8},
  pages={1735-1780},
  doi={10.1162/neco.1997.9.8.1735}
}

@article{processtransformer,
  author       = {Zaharah Allah Bukhsh and
                  Aaqib Saeed and
                  Remco M. Dijkman},
  title        = {ProcessTransformer: Predictive Business Process Monitoring with Transformer
                  Network},
  journal      = {CoRR},
  volume       = {abs/2104.00721},
  year         = {2021},
  url          = {https://arxiv.org/abs/2104.00721},
  eprinttype    = {arXiv},
  eprint       = {2104.00721},
  bibsource    = {dblp computer science bibliography, https://dblp.org}
}

@article{transformer,
  author       = {Ashish Vaswani and
                  Noam Shazeer and
                  Niki Parmar and
                  Jakob Uszkoreit and
                  Llion Jones and
                  Aidan N. Gomez and
                  Lukasz Kaiser and
                  Illia Polosukhin},
  title        = {Attention Is All You Need},
  journal      = {CoRR},
  volume       = {abs/1706.03762},
  year         = {2017},
  url          = {http://arxiv.org/abs/1706.03762},
  eprinttype    = {arXiv},
  eprint       = {1706.03762},
  bibsource    = {dblp computer science bibliography, https://dblp.org}
}

@article{kapoor2023reproducibilitycrisis,
  title={Leakage and the reproducibility crisis in machine-learning-based science},
  author={Kapoor, Sayash and Narayanan, Arvind},
  journal={Patterns},
  volume={4},
  number={9},
  year={2023},
  publisher={Elsevier}
}

@article{baker2016reproducibility,
  title={Reproducibility crisis},
  author={Baker, Monya},
  journal={nature},
  volume={533},
  number={26},
  pages={353--66},
  year={2016}
}

@inproceedings{wuyts2024sutran,
  title={Sutran: an encoder-decoder transformer for full-context-aware suffix prediction of business processes},
  author={Wuyts, Brecht and Vanden Broucke, Seppe and De Weerdt, Jochen},
  booktitle={2024 6th International Conference on Process Mining (ICPM)},
  pages={17--24},
  year={2024},
  organization={IEEE}
}

@inproceedings{brodersen2010balancedacc,
  title={The balanced accuracy and its posterior distribution},
  author={Brodersen, Kay Henning and Ong, Cheng Soon and Stephan, Klaas Enno and Buhmann, Joachim M},
  booktitle={2010 20th international conference on pattern recognition},
  pages={3121--3124},
  year={2010},
  organization={IEEE}
}

@inproceedings{spelmen2018review,
  title={A review on handling imbalanced data},
  author={Spelmen, Vimalraj S and Porkodi, R},
  booktitle={2018 international conference on current trends towards converging technologies (ICCTCT)},
  pages={1--11},
  year={2018},
  organization={IEEE}
}

@article{lecun2010mnist,
  title={MNIST handwritten digit database},
  author={LeCun, Yann and Cortes, Corinna and Burges, CJ},
  journal={ATT Labs [Online]. Available: http://yann.lecun.com/exdb/mnist},
  volume={2},
  year={2010}
}

@article{wmpvdaalst2012process,
  title={Process mining},
  author={Van Der Aalst, Wil},
  journal={Communications of the ACM},
  volume={55},
  number={8},
  pages={76--83},
  year={2012},
  publisher={ACM New York, NY, USA}
}

@inbook{hennig2025transformericpm,
  title = {Towards Accurate Predictions in ITSM: A Study on Transformer-Based Predictive Process Monitoring},
  ISBN = {9783031822254},
  ISSN = {1865-1356},
  url = {http://dx.doi.org/10.1007/978-3-031-82225-4_16},
  DOI = {10.1007/978-3-031-82225-4_16},
  booktitle = {Process Mining Workshops},
  publisher = {Springer Nature Switzerland},
  author = {Hennig,  Marc C.},
  year = {2025},
  pages = {214–226}
}

@inbook{hennig2025transformerbpm,
  title = {Leveraging Temporal Graphs for Enhancing Transformer-Based Predictive Process Monitoring},
  ISBN = {9783032028679},
  ISSN = {1611-3349},
  url = {http://dx.doi.org/10.1007/978-3-032-02867-9_18},
  DOI = {10.1007/978-3-032-02867-9_18},
  booktitle = {Business Process Management},
  publisher = {Springer Nature Switzerland},
  author = {Hennig,  Marc C. and Schmidt,  Rainer},
  year = {2025},
  month = aug,
  pages = {291–307}
}
%




\end{document}